\documentclass{article} 
\usepackage{iclr2025_conference,times}

\usepackage{amsmath,amsfonts,bm}

\def\eqref#1{equation~\ref{#1}}

\def\1{\bm{1}}

\DeclareMathAlphabet{\mathsfit}{\encodingdefault}{\sfdefault}{m}{sl}
\SetMathAlphabet{\mathsfit}{bold}{\encodingdefault}{\sfdefault}{bx}{n}

\usepackage{hyperref}
\usepackage{url}
\usepackage{graphicx}
\usepackage{wrapfig}
\usepackage{algorithm}
\usepackage{algorithmic}
\usepackage{booktabs}
\usepackage{multirow}

\title{Small Models are LLM Knowledge Triggers for Medical Tabular Prediction}

\author{Jiahuan Yan$^{1}$, Jintai Chen$^{2,3,}$\thanks{The corresponding author.}\ \ ,
Chaowen Hu$^1$, Bo Zheng$^1$, Yaojun Hu$^1$, \\\textbf{Jimeng Sun$^3$, Jian Wu$^{4,5}$} \\
$^1$College of Computer Science and Technology, Zhejiang University\\
$^2$Thrust of Artificial Intelligence, Information Hub, HKUST (GZ)\\
$^3$Computer Science Department, University of Illinois Urbana-Champaign\\
$^4$The Second Affiliated Hospital Zhejiang University School of Medicine\\
$^5$Zhejiang Key Laboratory of Medical Imaging Artificial Intelligence\\
\fontsize{9}{11}\selectfont
\texttt{\{jyansir,chaowenhu,zjuzhengbo,yaojunhu,wujian2000\}@zju.edu.cn}, \\
\fontsize{9}{11}\selectfont
\texttt{jimeng@illinois.edu}, \texttt{jintaichen@hkust-gz.edu.cn}
}

\iclrfinalcopy 
\begin{document}

\maketitle

\begin{abstract}
Recent development in large language models (LLMs) has demonstrated impressive domain proficiency on unstructured textual or multi-modal tasks. However, despite with intrinsic world knowledge, their application on structured tabular data prediction still lags behind, primarily due to the numerical insensitivity and modality discrepancy that brings a gap between LLM reasoning and statistical tabular learning. Unlike textual or vision data (e.g., electronic clinical notes or medical imaging data), tabular data is often presented in heterogeneous numerical values (e.g., CBC reports). This ubiquitous data format requires intensive expert annotation, and its numerical nature limits LLMs' capability to effectively transfer untapped domain expertise.
In this paper, we propose \texttt{SERSAL}, a general \textbf{self-prompting} method by synergy learning with small models to \textbf{enhance LLM tabular prediction in an unsupervised manner}. Specifically, \texttt{SERSAL} utilizes the LLM's prior outcomes as original soft noisy annotations, which are dynamically leveraged to teach a better small student model. Reversely, the outcomes from the trained small model are used to teach the LLM to further refine its real capability. This process can be repeatedly applied to gradually distill refined knowledge for continuous progress. Comprehensive experiments on widely used medical domain tabular datasets show that, without access to gold labels, applying \texttt{SERSAL} to OpenAI GPT reasoning process attains substantial improvement compared to linguistic prompting methods, which serves as an orthogonal direction for tabular LLM, and increasing prompting bonus is observed as more powerful LLMs appear. Codes are available at \url{https://github.com/jyansir/sersal}.
\end{abstract}

\section{Introduction}

The advancement of large language models (LLMs)~\citep{zhao2023survey} has made waves in both research and industry communities. Through friendly natural language interaction and powerful in-context reasoning ability, LLMs have shown their remarkable knowledge generalization to language processing~\citep{wei2021finetuned,wang2022language}, complex planning~\citep{qin2023toolllm,zan2023large} and even vertical domain (e.g., healthcare~\citep{cascella2023evaluating}, law~\citep{deroy2023ready}, chemistry~\citep{guo2023indeed}) tasks compared to past supervised pre-trained language models~\citep{kenton2019bert,radford2019language}, all achieved with suitable prompting and no fine-tuning, yet they are still struggling for the numeric tabular data.

\begin{figure*}[t]
\centering
\includegraphics[width=1.0\linewidth]{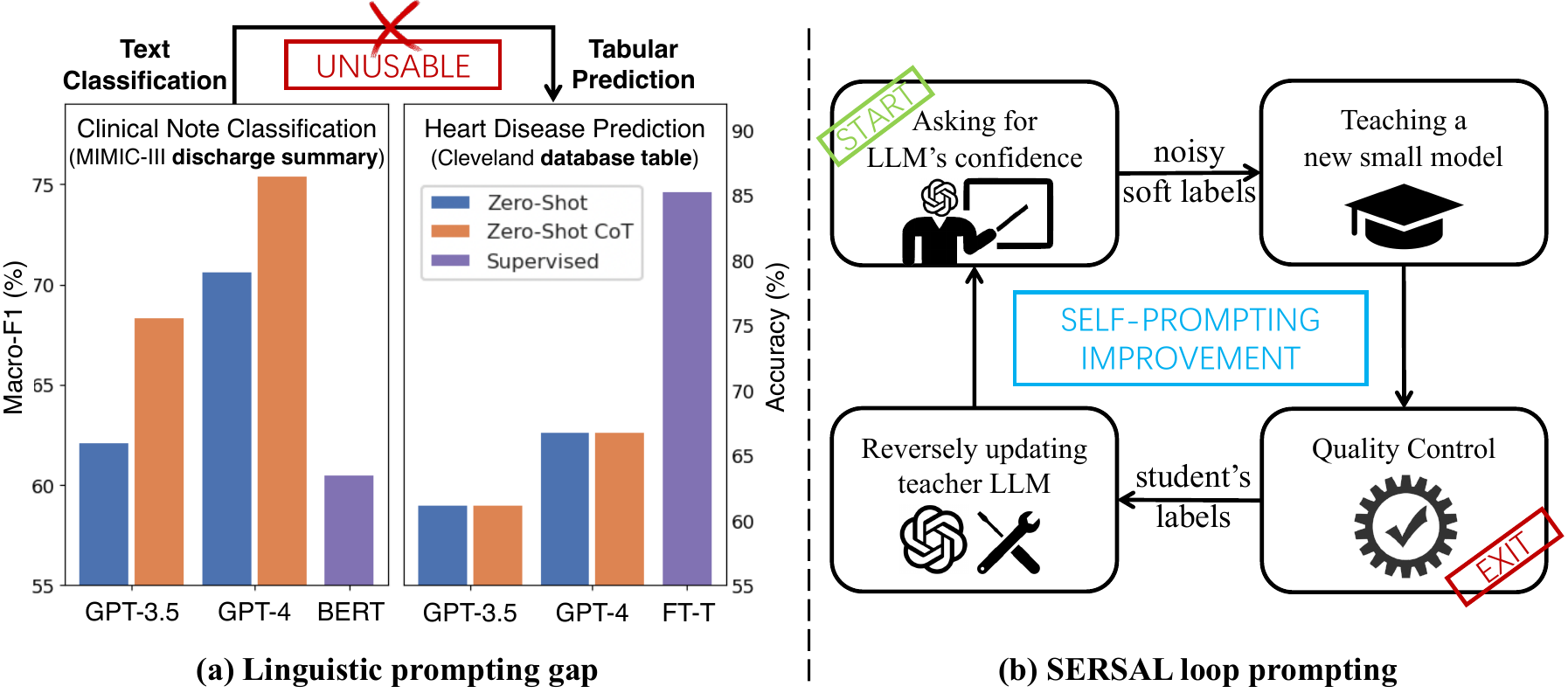}
\caption{(a) Comparison of prompting effectiveness on unstructured textual data~\citep{mullenbach2018explainable} and structured tabular data~\citep{detrano1989international} from medical domain, it is clearly seen, even with surprising medical expertise~\citep{nori2023can}, GPT-4 still struggles to catch up fully supervised small models (ClinicalBERT~\citep{huang2019clinicalbert} for textual tasks and FT-Transformer~\citep{gorishniy2021revisiting} for tabular ones) on tabular data, implying essential task discrepancy that makes it incompatible to rely on typical prompting techniques to unlock the potential of LLMs for tabular prediction. (b) Unsupervised \texttt{SERSAL} triggers LLM's knowledge using a small model.}
\label{main-fig}
\end{figure*}

For example, GPT-4 achieves 81.7 \% accuracy with zero-shot prompting on the United States Medical Licensing Examination (USMLE), which metric will be increased to 90.2 \% when meticulous prompts are designed~\citep{nori2023can}. In the left part of Fig.~\ref{main-fig}(a), our preliminary experiment exhibits performances of GPT-3.5, GPT-4 and the fully supervised BERT on top-5 ICD coding for MIMIC-III discharge summaries. Even with simple zero-shot prompting, GPT-3.5 has already surpassed the fine-tuned ClinicalBERT~\citep{huang2019clinicalbert} and can obtain further improvement with linguistic prompting tricks (e.g., zero-shot CoT~\citep{kojima2022large}). However, when handling medical tables with numerical value features, the trend becomes totally different, in the right part of Fig.~\ref{main-fig}(a), such significant prompting bonus disappears, suggesting an undeniable void in the current LLM prompting taxonomy tailored for tabular prediction. There are two key points causing the gap: 

\textbf{(i)} Existing competitions for general-purpose LLMs predominantly focus on their capabilities in processing unstructured data~\citep{zhang2024mm,zhang2024vision}, which is naturally different from structured tabular data characterized by dense heterogeneous numerical features~\citep{borisov2022deep,yan2023t2g,chen2022danets}, and the prevailing technical landscape of LLMs neglects nuanced sensitivity and understanding for numerical values~\citep{qian2023limitations,yan2024making}.

\textbf{(ii)} Most LLM tasks of interest can be formulated as sample-level data understanding then reasoning by generation, the input semantics are unstructured and detailed, while the tabular prediction (e.g., disease diagnosis with numerical metrics from medical examination and testing) requires overall statistical relation between numerical features and targets on the whole dataset or a specific task, which is difficult to access using a single tabular instance in high-level and constrained contexts.

Based on these observations, a straightforward question is, how to harness world knowledge of existing versatile LLMs, especially these commercial and blackbox (users cannot access the logit) ones~\citep{openai2022gpt3,openai2023gpt4}, to empower tabular prediction like disease diagnosis using medical testing results, which serves as a potential breakthrough for LLMs on statistical learning tasks.

To fill the aforementioned technical gap and extend LLM's capability to tabular prediction, we propose \texttt{SERSAL}, a \textbf{\underline{s}}yn\textbf{\underline{er}}gy learning pipeline between \textbf{\underline{s}}mall models \textbf{\underline{a}}nd \textbf{\underline{L}}LMs, requiring no gold labels. Different from existing prompting techniques designing hard or soft prompts to augment inputs for unstructured data understanding, our \texttt{SERSAL} contributes from a brand new perspective that \textbf{improves existing LLMs on statistical prediction for numeric tabular data}, providing an interface to adapt LLM untapped knowledge to such structured tabular data.
\texttt{SERSAL} helps a blackbox LLM trigger and refine its vertical capabilities for domain tabular data in an \textbf{unsupervised} manner with the following steps:
(1) Using simple zero-shot prompting to gather the LLM's output confidence as initial coarse annotations of the whole dataset;
(2) Teaching a better small tabular model (e.g., FT-Transformer~\citep{gorishniy2021revisiting}) from scratch based on the LLM's confidence like semi-supervised learning with noisy labels;
(3) Reversely fine-tuning the LLM using the outcomes of the trained small model to further update LLM's annotations in the next loop; The process from step (1) to (3) can be repeatedly applied to the LLM for continuous progress on specific tabular dataset. Essentially, \texttt{SERSAL} presents LLM prior knowledge on all tabular samples as ``indicators'' to a small model, which serves as a ``probe'' during learning correct patterns from the well-expressed clean part to feedback for LLM self-improvement.

In this paper, the main experiment is based on the well-known online blackbox LLMs, OpenAI GPT-3.5~\citep{openai2022gpt3} \& GPT-4~\citep{openai2023gpt4}, and as a prompting counterpart, our \texttt{SERSAL} can be directly transferred to other LLMs once the fine-tuning APIs are given. In a nutshell, our main contributions are:

\begin{itemize}
    \item For the first time, we bring the common challenge of existing general-purpose LLMs on numeric tabular prediction, a statistical learning featured task, to the spotlight that has not been covered by prevailing prompting techniques.
    \item We propose \texttt{SERSAL}, a novel unsupervised self-prompting method to adapt LLM's capability to tabular data prediction, which leverages synergy learning with small models to capture and feedback correct patterns from LLM intrinsic knowledge.
    \item Comprehensive experiments reveal that \texttt{SERSAL} is consistently more effective than common textual prompting methods on medical tabular datasets, with general feasibility in other vertical domains discussed.
\end{itemize}

\section{SERSAL: An LLM Self-Prompting Loop for Tabular Prediction}

We propose \texttt{SERSAL}, a synergy learning process using small models to trigger LLM's knowledge on tabular data, which is a fundamentally distinct prompting method and serves as a novel interface to extend current LLMs to tabular data prediction. Principally, \texttt{SERSAL} is inspired by the semi-supervised learning with noisy labels (LNL) and teacher-student training, while several key differences exist: (1) LNL setting requires a certain proportion of gold labels as the starting point, while \texttt{SERSAL} only access the LLM's \textbf{soft pseudo labels} (i.e., per-sample confidence) on the whole dataset; (2) In teacher-student paradigm the student model is primarily considered to be comparable to the teacher, while \texttt{SERSAL} conservatively \textbf{teaches a better student model} by dynamically learning from the relatively clean LLM's outputs and regularizing on noisy ones to avoid misleading confirmation bias~\citep{tarvainen2017mean}, which produces a better small model on the target task to form a co-teaching manner. The overall framework of \texttt{SERSAL} is outlined in Fig.~\ref{main-fig}(b) and formulated in Algorithm~\ref{alg:sersal}. Each part is detailed in the following subsections.

\subsection{Soft LLM Pseudo Labeling} \label{fine-grained-ann}

To access the prior knowledge of the LLM on a specific tabular dataset, we first query its confidence on each sample using simple zero-shot prompt template. Specifically, the prompt consists of a task description and listed feature specifications, for example, ``You are a professional doctor, here are some clinical metrics of a patient, please give a likelihood between 0 to 1 of suffering from a heart disease: [Age] 47 (years old); [Gender] Male; [Systolic Blood Pressure] 138 (mmHg); [Blood Lipid] 240 (mg/dL); \dots''. In this way the LLM's per-sample confidence on the whole dataset is gathered, though the initial zero-shot performance is often far away from the one of a supervised small tabular model (see Fig.~\ref{main-fig}(a) and Table~\ref{main-res}), we can dig into such fine-grained LLM confidence by separately judging then using underlying clean and noisy supervision signal to teach a robust small model.

\subsection{Small Model Teaching with Noisy LLM's Labels} \label{teaching}

This step aims to teach a better small model with the collected soft outputs from the LLM. Intuitively, such LLM confidence is a kind of noisy labels, thereby a straightforward insight is to reformulate the teaching process as learning with noisy labels (LNL). To sufficiently exploit LLM's prior knowledge, we adopt a ``semi-supervised'' learning strategy after dividing training samples into a more reliable labeled set and another unlabeled set, then the small model is fitted with the soft LLM's labels in the labeled set and regularized on the ones of the unlabeled set, the data partition is based on per-sample loss since deep neural networks tend to fit samples with clean labels faster than one with wrong labels according to the LNL theory~\citep{arpit2017closer}, thus lower loss often indicates relatively cleaner labels~\citep{chen2019understanding}.

In implementation, we use an adapted version of DivideMix~\citep{li2019dividemix}, a common semi-supervised LNL algorithm for image classification that dynamically fits a Gaussian Mixture Model on per-sample losses to distinguish probably clean and noisy LLM's labels and trains a pair of neural networks simultaneously to keep them diverged to avoid confirmation bias in single-model self-training~\citep{tarvainen2017mean}. Apart from adapting DivideMix to tabular data prediction, the used soft labels naturally apply label smoothing guided by the LLM. Besides, we leverage the pseudo labels with extreme confidence for early stopping with underlying assumption that annotations with extreme LLM's confidence is tend to be more accurate, which is observed in Fig.~\ref{conf-perform} and Fig.~\ref{conf-perform2}, and discussed in Sec.~\ref{exp:abl}. Specifically, we divide a training subset as the early stopping set $D_{\rm es} = \left \{ (\mathbf{X}_i, \bar{\mathbf{y}}_i) | {\rm max}(\hat{\mathbf{y}}_i) \ge \tau  \right \}$ to perform early stopping and hyper-parameter selection for the teaching process, where for the $i$-th sample, $\hat{\mathbf{y}}_i$ is its LLM's confidence vector, and $\bar{\mathbf{y}}_i$ is the corresponding hard labels (i.e., $\bar{\mathbf{y}}_i = {\rm argmax}(\hat{\mathbf{y}}_i)$), samples with maximum label confidence larger than threshold $\tau$ (we uniformly set $\tau = 0.9$ in the experiment) are considered to be accurate enough for early stopping. During ``semi-supervised'' teaching, samples in the early stopping set are also used since some domain (e.g., medicine) tabular datasets suffer from data inadequacy, and the reduction on training subsets may distort data distribution. We formulate this step in the line 3-5 of Algorithm~\ref{alg:sersal} and conduct related ablations in Sec.~\ref{exp:abl}.

In summary, this teaching step adopts semi-supervised LNL process to aggregate and distill prior knowledge into a small model to extend the LLM's real capabilities to tabular data prediction.

\subsection{Quality Control} \label{quality-ctr}

Since \texttt{SERSAL} operates iteratively, it requires a termination mechanism to control the loop exit. Here we provide three heuristic strategies, users can also define their own control flow in practice.
\begin{itemize}
    \item \textbf{Metric-based Control.} In Sec.~\ref{teaching} we define the high-confidence training subset as the early stopping set $D_{\rm es}$ which pseudo labels are relatively more accurate (see Fig.~\ref{conf-perform}). Therefore, users can inspect metrics (e.g., AUC or accuracy scores for classification) by treating these pseudo labels as the ''ground truth'' to control whether to end the loop.
    \item \textbf{External Validation Control.} If budget permits, human experts can collect and annotate appropriate external data as a validation set, e.g., in hospitals, regular medical data quality inspection needs to sample and label a small part of data, and learning quality can be assessed with such external labeled set.
    \item \textbf{Rule-based Control.} For example, users can define a fixed iteration time.
\end{itemize}

For simplicity, in the main experiment we uniformly report one-loop \texttt{SERSAL} performances in medical and other domain datasets (Table~\ref{main-res} \&~\ref{other-domain}), which has significantly surpassed the ones of prevailing prompting methods, and further discuss the effectiveness of multi-loop \texttt{SERSAL} in Sec.~\ref{exp:further-loop}.

\subsection{Reverse LLM Tuning} \label{llm-tuning}

The final step is to reversely teach the LLM using the well-trained small model to feedback the aggregated knowledge. Similar to using LLM's soft confidence to teach the small model in Sec.~\ref{teaching}, we also use soft confidence from the small model to fine-tune the LLM (fine-tune the online blackbox GPTs through their APIs in experiment). Specifically, the training samples are re-labeled by the small model with its guessed probabilities (line 7-8 in Algorithm~\ref{alg:sersal}), the same prompt templates in Sec.~\ref{fine-grained-ann} are used to construct the training corpus for the LLM. To avoid the excessive memorization of the LLM on the small model outputs~\citep{bordt2023elephants}, we employ a conservative tuning strategy that sets the maximum training epoch to 3 with proper early stopping (the fine-tuning APIs of GPT-3.5 \& GPT-4 provide automatic early stopping in default), making the LLM slightly fitted on the guessed labels while keeping a non-zero minimum training loss. Then the updated LLM initiates the next \texttt{SERSAL} loop, forming an iterative process.

\begin{algorithm}[tb]
    \caption{Unsupervised \texttt{SERSAL}. Line 2: LLM pseudo labeling (Sec.~\ref{fine-grained-ann}); Line 3-5: Small model teaching (Sec.~\ref{teaching}); Line 6: Quality control (Sec.~\ref{quality-ctr}); Line 7-9: Reverse tuning (Sec.~\ref{llm-tuning}).}
    \label{alg:sersal}
    \textbf{Input}: Unlabeled training set $\textbf{X}_{\rm train}$ and test set $\textbf{X}_{\rm test}$, large language model $f_{\rm LLM}^{(0)}$\\
    \textbf{Parameter}: Confidence threshold $\tau$, quality control function $f_{\rm ctr}$\\
    \textbf{Output}: Improved zero-shot tabular prediction $\textbf{y}_{\rm test}^\ast$
    \begin{algorithmic}[1] 
        \STATE Let $t=1$. {// Initialize iteration number}
        \STATE {Softly labeled dataset} $D_{\rm train}^{(t)}=(\mathbf{X}_{\rm train}, \hat{\mathbf{y}}^{(t)})$ {by current} $f_{\rm LLM}^{(t)}$.
        \STATE {Randomly initialize a small tabular model} $\theta^{(t)}$.
        \STATE {Select early stopping set} $D_{\rm es}^{(t)} = \left \{ (\mathbf{X}_i, \bar{\mathbf{y}}_i^{(t)}) | {\rm max}(\hat{\mathbf{y}}_i^{(t)}) \ge \tau \right \} \subseteq D_{\rm train}^{(t)}$.
        \STATE $\theta^{\ast(t)}$ = {DivideMix}$(D_{\rm train}^{(t)}, D_{\rm es}^{(t)}, \theta^{(t)}, \tau)$. {// Adapted DivideMix~\citep{li2019dividemix}}
        \WHILE{$f_{\rm ctr}(\theta^{\ast(t)},\textbf{X})$}
        \STATE $\textbf{y}_{\rm sm}^{(t)}={\rm Predict}(\textbf{X}_{\rm train}; \theta^{\ast(t)})$. {// Soft label guessing by the small model}
        \STATE $\hat{\textbf{y}}_{\rm sm}^{(t)}={\rm Sharpen}(\textbf{y}_{\rm sm}^{(t)},{\rm temperature=0.1})$. {// Simple temperature sharpening}
        \STATE $f_{\rm LLM}^{(t+1)}={\rm Finetune}(\textbf{X}_{\rm train},\hat{\textbf{y}}_{\rm sm}^{(t)}, f_{\rm LLM}^{(t)})$. {// Reversely tune the LLM with guessed labels}
        \STATE $t=t+1$.
        \STATE Repeat Line 2-5. {// Self-prompting loop}
        \ENDWHILE
        \STATE $\textbf{y}_{\rm test}^\ast = {\rm Predict}(\textbf{X}_{\rm test};\theta^{\ast(t)})$. {// Final prediction with the taught small model}
        \STATE \textbf{return} $\textbf{y}_{\rm test}^\ast$
    \end{algorithmic}
\end{algorithm}

\section{Experiments}

In this section, we first compare \texttt{SERSAL} with prevailing prompting techniques (using GPT-3.5 \& GPT-4) and the fully supervised small tabular models on extensive medical tabular datasets in Sec.~\ref{sec:main}. Next, we conduct ablation on several key adaptations in semi-supervised learning with noisy labels (LNL) in Sec.~\ref{teaching} and inspect the effectiveness of multi-loop \texttt{SERSAL} in Sec.~\ref{exp:further-loop}. Also, we discuss the general adaptability of \texttt{SERSAL} on tabular data from other non-medical domains in Sec.~\ref{sec:other-domain}. Besides, we explore the method interpretability by visualizing Shapely Value variation during \texttt{SERSAL} process in Sec.~\ref{exp:interpretability}.

\subsection{Experimental Setup} \label{exp:setup}

\paragraph{Datasets}

We evaluate on ten widely recognized medical diagnosis tabular datasets on various diseases: Heart Failure Prediction (HF, ~\cite{detrano1989international}), Lung Cancer Prediction (LC, ~\cite{ahmad2020new}), Early Classification of Diabetes (ECD, ~\cite{islam2020likelihood}), Indian Liver Patient Records (LI, ~\cite{ramana2012critical}), Hepatitis C Prediction (HE, ~\cite{hoffmann2018using}), Pima Indians Diabetes Database (PID, ~\cite{smith1988using}), Framingham Heart Study (FH, ~\cite{o2008cardiovascular}), Stroke Prediction (ST, ~\cite{fedesoriano2020stroke}), COVID-19 Presence(CO, ~\cite{hemanthhari2020symptom}) and Anemia Disease (AN, ~\cite{kilicarslan2021hybrid}). Besides, datasets in clinical trail~\citep{wang2022transtab} and open domains~\citep{gorishniy2021revisiting} are added to further inspect the effectiveness of \texttt{SERSAL} in difficult tasks and general data domains respectively. We split each tabular dataset (80 \% for training and 20 \% for testing), and keep the same label distribution in each split. Statistics of medical diagnosis datasets are given in Table~\ref{data-stat}. All evaluated datasets are binary classification tasks.

\begin{table*}[ht]
\centering
\resizebox{\textwidth}{!}{
\begin{tabular}{lcccccccccc}
\toprule
Dataset& HF& LC& ECD& LI& HE& PID& FH& ST& CO& AN\\
\midrule
\# features& 13& 15& 16& 10& 12& 8& 15& 7& 20& 24\\
\# samples& 303& 309& 520& 583& 615& 768& 4238& 5110& 5434& 15300\\
P/N& 0.80& 6.92& 1.60& 2.51& 0.11& 0.54& 0.18& 0.04& 4.17& 0.57\\
disease& Heart& Lung& Diabetes& Liver& Hepatitis C& Diabetes& Heart& Stroke& COVID-19& Anemia\\
\bottomrule
\end{tabular}}
\caption{Dataset statistics of ten medical diagnosis datasets for binary classification on various diseases. ``P/N'' denotes the amount ratio of positive samples and negative ones.}\label{data-stat}
\end{table*}

\paragraph{Compared Methods}

Since \texttt{SERSAL} serves as an unsupervised self-prompting method for LLM tabular prediction, we compare with existing linguistic prompting methods for LLM usage in general textual and tabular tasks, which focus on meticulously designed prompt texts: (1) \textbf{Zero-Shot Prompting} (0-shot) is the straightforward prompt that contains no examples; (2) \textbf{Zero-Shot CoT Prompting}~\citep{kojima2022large} (CoT) is a popular prompting method which asks the LLMs to answer with intermediate reasoning steps to enable complex reasoning capabilities; (3) \textbf{8-shot Prompting} (8-shot) is a common few-shot prompt setting in standard prompting studies~\citep{wei2022chain,kojima2022large,nori2023can}, it provides eight labeled samples (exemplars) to enrich prompt contexts and steer the LLM to the better outputs, in the experiment we randomly sample eight training examples and control the same positive-negative ratio (i.e., ``P/N'' in Table~\ref{data-stat}) with at least one example for each class; (4) \textbf{TabLLM}~\citep{hegselmann2023tabllm} and (5) \textbf{LIFT}~\citep{dinh2022lift} are two recent known linguistic prompt schemes for textualizing tabular data to fine-tune LLMs with gold labels, though TabLLM was additionally evaluated in zero-shot settings, \textbf{none of them are originally proposed for unsupervised tabular scenarios}, here we use their zero-shot schemes for comparison. Additionally, we provide a \textbf{fully supervised small tabular model} (FSSM) group using FT-Transformer~\citep{gorishniy2021revisiting} for reference representing traditional supervised learning paradigm by fine-tuning dataset-specific small models. 

\paragraph{Implementation Details}

All experiments are conducted with PyTorch on Python 3.8 and run on NVIDIA RTX 3090. For the small model, we uniformly use FT-Transformer with the default model and training configurations provided in the original paper~\citep{gorishniy2021revisiting}. For \texttt{SERSAL}, the only adjustable hyper-parameter is the temperature of DivideMix~\citep{li2019dividemix} with choices of 0.5, 5.0 and 10.0 in line 5 of Algorithm~\ref{alg:sersal}, which is selected by the metric of the early stopping set ($D_{\rm es}^{(t)}$ in line 4 of Algorithm~\ref{alg:sersal}). The LLMs in the experiment includes OpenAI GPT-3.5 \& GPT-4 to inspect the effectiveness of \texttt{SERSAL} across different LLM capabilities.

\subsection{Why We Need SERSAL?} \label{sec:main}

\begin{table*}[th]
\centering
\resizebox{\textwidth}{!}{
\begin{tabular}{@{}l|cccccccccc@{}}
\toprule
       & HF    & LC    & ECD   & LI    & HE     & PID   & FH    & ST    & CO    & AN    \\ \midrule
Random guessing & 37.22 & 40.18 & 46.25 & 50.28 & 62.73  & 63.24 & 50.39 & 41.76 & 71.55 & 51.28 \\
FSSM$^\ast$(supervised FT-T)   & 88.19 & 86.61 & 99.60 & 78.94 & 100.00 & 84.72 & 66.25 & 82.98 & 99.91 & 99.92 \\ \midrule
0-shot (GPT-3.5) & 71.88 & 78.87 & 85.71 & 76.81 & 68.51  & 73.12 & 60.32 & 63.01 & 82.60 & 90.43 \\
8-shot$^\ast$ (GPT-3.5) & 73.65 & 78.87 & \textbf{87.68} & 76.81 & 68.51  & 73.12 & 58.27 & 60.85 & 77.63 & 87.19 \\
CoT (GPT-3.5) & 71.88 & 78.87 & 82.36 & 76.81 & 68.51  & 70.83 & 60.32 & 63.01 & 82.60 & 90.43 \\
TabLLM (GPT-3.5) & 76.37 & 78.87 & 87.06 & 78.24 & 74.39 & 75.69 & 61.78 & 68.48 & 85.78 & 89.11 \\
LIFT (GPT-3.5) & 78.23 & 80.69 & 83.92 & 73.60 & 72.57 & 73.12 & 60.32 & 70.92 & 87.93 & 90.43 \\
\texttt{SERSAL} (GPT-3.5) & \textbf{91.39} & \textbf{85.42} & 86.40 & \textbf{79.39} & \textbf{85.14}  & \textbf{78.97} & \textbf{63.97} & \textbf{76.36} & \textbf{96.85} & \textbf{98.37} \\ \midrule
TabLLM+\texttt{SERSAL} (GPT-3.5) & 93.82 & 85.42 & 88.39 & 80.71 & 89.27 & 82.54 & 65.02 & 81.74 & 97.51 & 98.16 \\
\texttt{SERSAL} (GPT-4) & 94.18 & 86.93 & 92.68 & 82.51 & 92.76 & 82.39 & 67.14 & 81.23 & 97.96 & 98.82 \\ \bottomrule
\end{tabular}}
\caption{The AUC scores (\%) of different tabular prediction schemes on 10 medical diagnosis datasets. The top part is the traditional supervised small models, the middle one is compared LLM prompting methods (the top performances are marked in \textbf{bold}), the bottom part is additional combinations. Here the results of \texttt{SERSAL} are only based on a single loop. ``$\ast$'' denotes the groups use gold labels. ``FSSM'' is the fully supervised FT-Transformer. The results on more difficult clinical trial datasets are given in Table~\ref{clinical-trial}.}
\label{main-res}
\end{table*}

\paragraph{Main Results Analysis}

The performances of different LLM prompting baselines are reported in the middle part of Table~\ref{main-res}. An overall trend is that, when the GPT-3.5 meets medical domain tabular prediction tasks, the results using common prompting methods are consistently better than the ones of random guessing, demonstrating the general-purpose LLMs indeed contain medical domain expertise inherently, but they are still far from the traditional supervised small models (see group ``FSSM''), and further performance enhancement can not be achieved through usual prompting tricks as in textual tasks (see Fig.~\ref{main-fig}(a)). Specifically, we observe 8-shot prompting slightly benefits the performances in small-scale datasets (e.g., HF and ECD) but hurts in the larger datasets (e.g., FH, ST, CO and AN) compared to the 0-shot prompting, which may be explained by the representativeness of the used examples, since the distribution of the smaller datasets are more likely to be covered by few examples, thus 8-shot performs better as data scale decreases, and vice versa. For 0-shot CoT prompting, it does not affect the overall results in most cases, but we find slight performance decline in two diabetes datasets (i.e., ECD and PID), this may be caused by the over-consideration of CoT on noisy features since diabetes can be diagnosed with several prominent features (e.g., blood sugar and lipid). Although carefully crafted prompt templates from recent LLM in-context tabular learning studies (i.e., TabLLM and LIFT) show modest improvement, \textbf{they still follow the linguistic nature to process numeric tabular data}, and are primarily designed for LLM in-context few-shot learning or supervised fine-tuning. Our \texttt{SERSAL} \textbf{explores a fundamentally novel prompting mechanism exploiting the information gain in the LLM's noisy outputs}, which breaks through the predicament from an orthogonal perspective and serves as an interface to effectively adapt the LLM's domain knowledge to numeric tabular data. After applying \texttt{SERSAL}, without access to gold labels, the GPT-3.5 is able to achieve significantly better reasoning on these medical domain tasks, with many cases competitive with the supervised small models.

\paragraph{Orthogonal Technical Contribution}

Based on the above analysis, \texttt{SERSAL} works in a distinct underlying mechanism, and we can jointly adopt \texttt{SERSAL} and previous linguistic prompting methods for better combined performances (see group ``TabLLM+\texttt{SERSAL}'' in Table~\ref{main-res}).

\paragraph{Continuous Performance Growth}

We additionally apply \texttt{SERSAL} to OpenAI GPT-4 on medical diagnosis datasets (the bottom part of Table~\ref{main-res}) and more difficult clinical trial datasets (see Table~\ref{clinical-trial}). It can be seen \texttt{SERSAL} can further realize substantial performance gains as the capability of used LLMs becomes more powerful, which can even surpass the traditional supervised paradigm (N00041119 and N00312208 datasets in Table~\ref{clinical-trial}), indicating ample room for continuous prompting bonus in \texttt{SERSAL} alongside the emergence of more advanced LLMs.

\subsection{Several Key Adaptations} \label{exp:abl}

\begin{table*}[h]
\centering
\resizebox{\textwidth}{!}{
\begin{tabular}{@{}l|cccccccccc@{}}
\toprule
            & HF    & LC    & ECD   & LI    & HE    & PID   & FH    & ST    & CO    & AN    \\ \midrule
\texttt{SERSAL}      & 91.39 & 85.42 & 86.40  & 79.39 & 85.14 & 78.97 & 63.97 & 76.36 & 96.85 & 98.37 \\ \midrule
w/o soft pseudo & 84.58 & 76.58 & 87.24 & 78.25 & 75.79 & 75.93 & 62.58 & 75.05 & 93.97 & 97.53 \\
w/o ES      & 84.03 & 74.11 & 75.92 & 59.39 & 47.41 & 68.43 & 57.08 & 74.70  & 90.57 & 97.57 \\ \bottomrule
\end{tabular}}
\caption{The AUC scores of ablation on two key adaptations. ``w/o soft pseudo'' means replacing the LLM's soft outputs with hard ones  during teaching the student model, ``w/o ES'' denotes no early stopping during DivideMix (line 5 in Algorithm~\ref{alg:sersal}).}
\label{tab:abl}
\end{table*}

\begin{wrapfigure}{r}{0pt}
\includegraphics[width=0.5\textwidth]{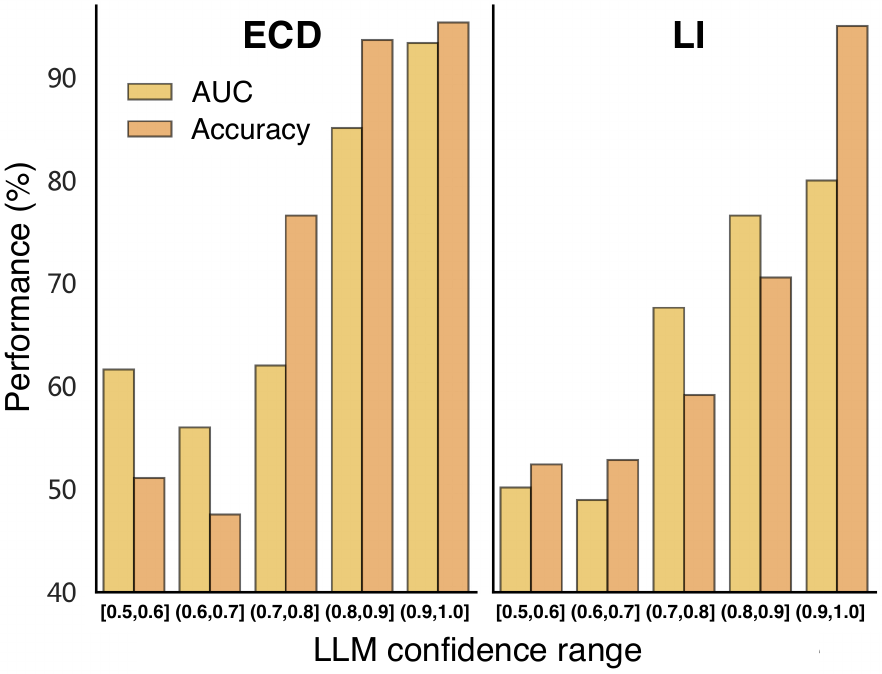}
\caption{Performances in different LLM's confidence ranges on ECD and LI datasets. Extreme-confidence samples are relatively more reliable.}
\vskip -1.5 em
\label{conf-perform}
\end{wrapfigure}

In Sec.~\ref{teaching}, to adapt the LLM's outputs to a semi-supervised LNL process to teach a small model, we gather per-sample confidence from the LLM as soft noisy annotations and heuristically select samples with extreme confidence for early stopping. In this section, we will analyze the effect of the two designs which distinguish our \texttt{SERSAL} from traditional LNL settings.

\paragraph{The Effect of using Soft Labels}

We query soft confidence from the LLM (see Sec.~\ref{fine-grained-ann}) rather than directly using hard outputs for small model teaching. The prediction probabilities inherently reflect the LLM's prior knowledge as well as uncertainty on the domain tabular data and can be naturally treated as a kind of label smoothing. Besides, the probability values can be used to select relatively reliable labels to early stop the teaching process and avoid overfitting.
In Table~\ref{tab:abl} we compare the effect of using soft labels by replacing it with hard ones during \texttt{SERSAL} reasoning (group ``w/o soft pseudo''). We find that using hard ones is usually suboptimal since it loses both prediction uncertainty and label smoothing, which is unable to exploit fine-grained LLM's knowledge.

\paragraph{The Effect of Early Stopping}

In addition to using LLM's soft outputs, a relatively clean training subset is selected by threshold clipping on the per-sample confidence (line 4 in Algorithm~\ref{alg:sersal}) for early stopping. 
Table~\ref{tab:abl} report the ablation results by directly training 100 epochs (group ``w/o ES''). It can be clearly seen, simply following the original DivideMix is far from the desired results, since tabular features are heterogeneous and high-level compared to the well-patterned pixels of images~\citep{tabcaps,yan2024making}, and in medical tabular domain the typically limited available data further makes it prone to overfit without early stopping, for example, except large AN dataset, all other tabular datasets appear to be significantly impacted by removing the early stopping mechanism. The heuristic design of selecting extreme-confidence sample is inspired from the empirical assumption that confident predictions from the LLM are more likely to be accurate, which is supported by the performance variation of different confidence ranges in Fig.~\ref{conf-perform} and Fig.~\ref{conf-perform2}.

\subsection{Effectiveness of Multi-loop SERSAL} \label{exp:further-loop}

Since \texttt{SERSAL} can be iteratively applied to the LLM (see Fig.~\ref{main-fig}(b)), we further inspect the effectiveness of multi-loop \texttt{SERSAL} for GPT-3.5 reasoning. Specifically, we repeat the pipeline three times on ECD and LI datasets, the result variations are reported in Table~\ref{sersal-iter}.

\begin{wraptable}{r}{0pt}
\centering
\resizebox{0.5\textwidth}{!}{
\begin{tabular}{@{}c|cc|cc@{}}
\toprule
\multirow{2}{*}{\# Loop} & \multicolumn{2}{c|}{ECD}                      & \multicolumn{2}{c}{LI}                    \\ \cmidrule(l){2-5} 
                           & \multicolumn{1}{c|}{\texttt{SERSAL}}     & LLM 0-shot    & \multicolumn{1}{c|}{\texttt{SERSAL}}    & LLM 0-shot   \\ \midrule
1                          & \multicolumn{1}{c|}{86.40} & 85.71  & \multicolumn{1}{c|}{79.39} & 76.81 \\
2                          & \multicolumn{1}{c|}{87.00} & 86.42 & \multicolumn{1}{c|}{82.47} & 80.26 \\
3                          & \multicolumn{1}{c|}{89.00}        & 87.81        & \multicolumn{1}{c|}{84.07} & 82.91 \\ \bottomrule
\end{tabular}}
\caption{The AUC score variation of \texttt{SERSAL} outputs and zero-shot prompting of the tuned GPT-3.5 (LLM 0-shot) on LI and ECD datasets during three loops. ``\# Loop'' is the same as the variable $t$ in line 1 of of Algorithm~\ref{alg:sersal}. LLM 0-shot group at the first loop is the original LLM.}
\label{sersal-iter}
\end{wraptable}

During three loops, progressive improvement on both the small model (\texttt{SERSAL} outputs are from the well-trained small model of each loop) and the GPT-3.5 is observed. Surprisingly, even inferior to the 8-shot prompting baseline on ECD dataset after the first loop (see Table~\ref{main-res}), we find \texttt{SERSAL} can reduce the gap and even outperform few-shot baselines after several loops. Such continuous progress probably comes from the synergy learning between the small model and the LLM that \textbf{shares a similar underlying mechanism of co-teaching}~\citep{han2018co}, i.e., both sides dynamically learn from a part of reliable pseudo labels from each other and it makes them diverged to avoid confirmation bias, forming a mutual improvement manner to aggregate and refine LLM's untapped domain knowledge for tabular prediction.

\subsection{Gneralized Data Adaptability on Other Domains} \label{sec:other-domain}

\begin{wraptable}{r}{0pt}
\centering
\resizebox{0.5\textwidth}{!}{
\begin{tabular}{@{}l|cccc@{}}
\toprule
         & Churn    & Credit  & Adult     & Fake  \\ \midrule
domain   & Business & Finance & Sociology & N/A   \\
\# features & 10       & 10      & 14        & 6     \\
\# samples  & 10000    & 16714   & 48842     & 1000  \\ \midrule
Random guessing   & 66.35    & 43.80   & 58.73     & 53.85 \\
FSSM$^\ast$     & 86.27    & 84.88   & 91.39     & 55.31 \\ \midrule
0-shot (GPT-3.5)     & 77.81    & 69.05   & 75.10     & 46.28 \\
\texttt{SERSAL} (GPT-3.5)   & 83.29    & 79.36   & 88.72     & 38.72 \\ \bottomrule
\end{tabular}}
\caption{The dataset statistics and AUC scores on other non-medical domains. ``Fake'' is a generated dataset with random labels and features. The denotations follow the ones in Table~\ref{data-stat} and Table~\ref{main-res}.}
\label{other-domain}
\end{wraptable}

In this section, we further explore the data adaptability of \texttt{SERSAL} on other non-medical domains. We evaluate on three classic binary classification datasets: Churn Modeling~\citep{churn}, Credit~\citep{credit} and Adult~\citep{kohavi1996scaling}, which are widely included in general tabular prediction studies~\citep{gorishniy2021revisiting,yan2023t2g,grinsztajn2022tree}. Additionally, we build a dataset ``Fake'' by randomly generating features and binary labels to emulate an extreme case where the LLM has no relevant knowledge at all. The data information and the results are given in Table~\ref{other-domain}. As in the medical domain, the GPT-3.5 indeed holds the world knowledge and can directly achieve the considerable results with simple zero-shot prompting, and \texttt{SERSAL} further enhances the performance significantly. However, when facing the tabular data from an unknown domain (i.e., the Fake dataset), the LLM outputs high confidence on wrong labels, \texttt{SERSAL} is unable to recognize such totally misleading bias. Therefore, our \texttt{SERSAL} shares the same basic limitation as other linguistic prompting methods that the applied LLMs require a certain level of knowledge in the target domain.

\subsection{Interpretability of SERSAL} \label{exp:interpretability}

\begin{figure*}[ht]
\centering
\includegraphics[width=1.0\linewidth]{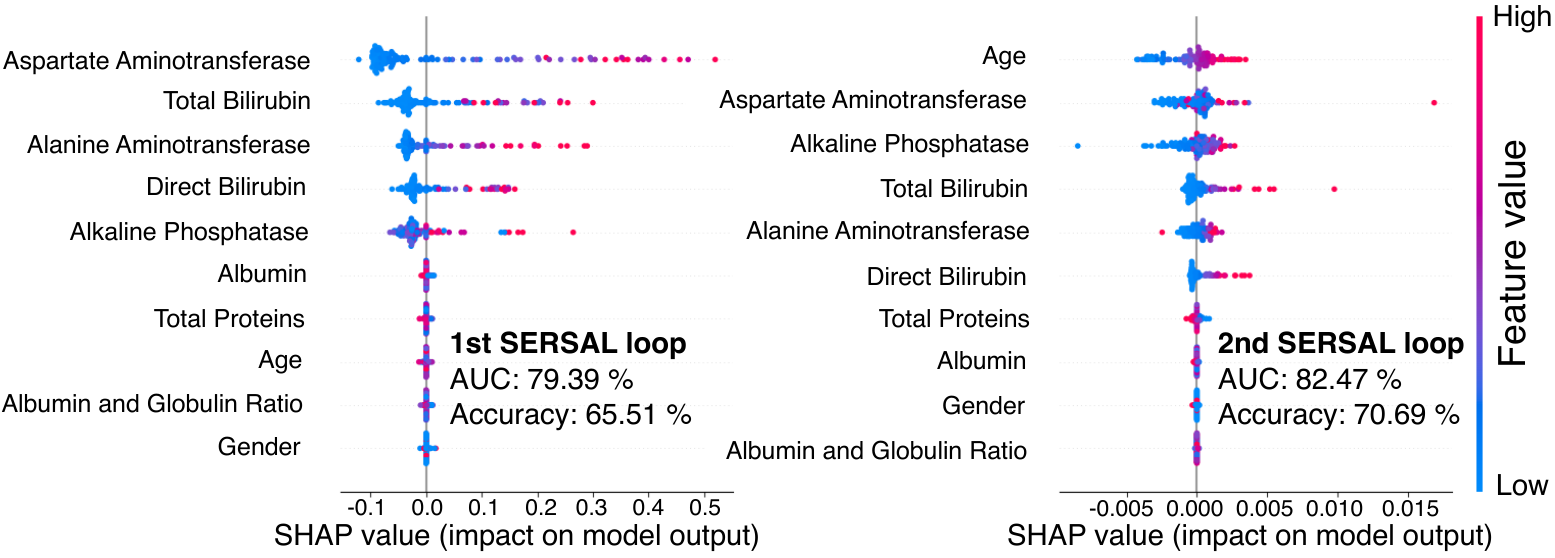}
\caption{Interpretability visualization from feature importance perspective: the variation of the Shapley Values (treat \texttt{SERSAL} outputs as the targets) and performances on Indian Liver Patient Records (LI dataset) after one and two \texttt{SERSAL} loops using GPT-3.5.}
\label{shap-liver}
\end{figure*}

In Fig.~\ref{shap-liver} we visualize the variation of Shapely Values on Indian Liver Patient Records (LI) dataset after one (left) and two (right) \texttt{SERSAL} loops by treating the predictions (i.e., Algorithm~\ref{alg:sersal}) as targets. It can be clearly seen the feature ``Age'' is adequately considered after one loop self-prompting, which highlights a strong and reasonable positive correlation between age and the incidence of liver diseases that aligns with the medical expertise. Besides, a negative correlation with ``Total Proteins'', a guiding clinical metric reflecting the liver's synthetic function, is captured in the right figure to contribute the prediction, since a lower total protein level indicates a risk of liver cirrhosis. These two reasonable changes of feature importance interpret the \texttt{SERSAL} prompting is able to iteratively refine the domain expertise in the LLM, calibrating the statistical feature-target relationship for better reasoning results during the process.

\section{Related Work}

\paragraph{Prompt Engineering for In-Context Learning}

Prompt engineering is a flourishing discipline for better LLM reasoning through meticulously designed linguistic input contexts or interaction process. The most common and straightforward prompting is the single-round instruction that directly asks with zero (zero-shot) or several (few-shot) demonstrations~\citep{brown2020language,wei2021finetuned}, but such prompt style fails to work in more complex reasoning tasks~\citep{wei2022chain}. To tackle this deficiency and improve the LLM's capacity on a wide range of tasks, recently, studies on more advanced prompting methods are emerging, such as chain-of-thought (CoT)~\citep{wei2022chain,kojima2022large,zhang2022automatic}, tree-of-thought (ToT)~\citep{yao2023tree} and self-consistency~\citep{wang2022self}. However, current prompting methods are mostly designed to serve unstructured data tasks~\citep{zhang2023multimodal}. Although recent studies on LLM in-context learning for tabular data (e.g., TabLLM~\citep{hegselmann2023tabllm}, LIFT~\citep{dinh2022lift}) propose table-friendly prompting strategies, their linguistic nature still hinders the numeric table understanding~\citep{yan2024making}.

\paragraph{Semi-supervised Learning with Noisy Labels}

Semi-supervised learning treats the unlabeled samples as regularization for better model generalization~\citep{lee2013pseudo,tarvainen2017mean,miyato2019virtual,berthelot2019mixmatch}. Recently, the related theory has been introduced to noisy label learning scenarios~\citep{song2022learning} that dynamically divide samples into clean labeled group and noisy unlabeled one~\citep{li2019dividemix} to achieve robust learning from noisy labels.

\paragraph{LLMs for Tabular Data Prediction}

As a machine learning task in tabular data applications, tabular prediction has gained increasing attention from the research community due to the heterogeneous nature and numerical features of tabular data compared to other unstructured modalities. Previous studies in tabular prediction models focus on designing tailored neural networks~\citep{arik2021tabnet,gorishniy2021revisiting,yan2024team,chen2024can} to emulate and surpass traditional tree-based models~\citep{chen2016xgboost,ke2017lightgbm} under fully supervised paradigm. More recently, motivated by the widespread success of pre-trained language models~\citep{brown2020language,openai2022gpt3,openai2023gpt4}, the unique bonus of neural networks is exploited in tabular model development, such as pre-training~\citep{wang2022transtab} and in-context learning~\cite{hollmann2022tabpfn}, and open-sourced LLMs have been popular base models to be adaptively pre-trained for better tabular prediction since their inherent knowledge~\citep{yan2024making,wen2024supervised}. However, current exploration on tabular prediction LLMs involves costly pre-training on large-scale tabular data, which requires access to LLM codes and parameters, and heavy adaptation to tabular data may impact the original usability of the LLMs on other unstructured data tasks.

\section{Conclusions}

This paper revealed the common challenge of existing general-purpose LLMs on tabular prediction and proposed \texttt{SERSAL}, a novel unsupervised self-prompting method in non-linguistic mechanism that triggers the LLM's domain knowledge for better tabular prediction. This is achieved through a co-teaching process between the LLM and a well-taught small tabular model which learn from the other's noisy outputs to aggregate and refine the LLM's untapped capabilities. Extensive experiments on medical and non-medical domain tabular datasets demonstrate that, as an orthogonal prompting landscape, \texttt{SERSAL} is consistently suitable for extending the potential of LLMs to numeric tabular data.


\subsubsection*{Acknowledgments}

This research was partially supported by National Natural Science Foundation of China under grant No. 12326612, Zhejiang Key R\&D Program of China under grant No. 2023C03053, and Zhejiang Key Laboratory of Medical Imaging Artificial Intelligence.

\bibliography{iclr2025_conference}

\begin{thebibliography}{66}
\providecommand{\natexlab}[1]{#1}
\providecommand{\url}[1]{\texttt{#1}}
\expandafter\ifx\csname urlstyle\endcsname\relax
  \providecommand{\doi}[1]{doi: #1}\else
  \providecommand{\doi}{doi: \begingroup \urlstyle{rm}\Url}\fi

\bibitem[Ahmad \& Mayya(2020)Ahmad and Mayya]{ahmad2020new}
Ahmad~S Ahmad and Ali~M Mayya.
\newblock A new tool to predict lung cancer based on risk factors.
\newblock \emph{Heliyon}, 2020.

\bibitem[Arik \& Pfister(2021)Arik and Pfister]{arik2021tabnet}
Sercan~{\"O} Arik and Tomas Pfister.
\newblock {TabNet}: Attentive interpretable tabular learning.
\newblock In \emph{AAAI}, 2021.

\bibitem[Arpit et~al.(2017)Arpit, Jastrzkebski, Ballas, Krueger, Bengio, Kanwal, Maharaj, Fischer, Courville, Bengio, et~al.]{arpit2017closer}
Devansh Arpit, Stanislaw Jastrzkebski, Nicolas Ballas, David Krueger, Emmanuel Bengio, Maxinder~S Kanwal, Tegan Maharaj, Asja Fischer, Aaron Courville, Yoshua Bengio, et~al.
\newblock A closer look at memorization in deep networks.
\newblock In \emph{ICML}, 2017.

\bibitem[Berthelot et~al.(2019)Berthelot, Carlini, Goodfellow, Papernot, Oliver, and Raffel]{berthelot2019mixmatch}
David Berthelot, Nicholas Carlini, Ian Goodfellow, Nicolas Papernot, Avital Oliver, and Colin~A Raffel.
\newblock Mixmatch: A holistic approach to semi-supervised learning.
\newblock In \emph{NeurIPS}, 2019.

\bibitem[Bordt et~al.(2023)Bordt, Nori, and Caruana]{bordt2023elephants}
Sebastian Bordt, Harsha Nori, and Rich Caruana.
\newblock Elephants never forget: Testing language models for memorization of tabular data.
\newblock In \emph{NeurIPS 2023 Second Table Representation Learning Workshop}, 2023.

\bibitem[Borisov et~al.(2022)Borisov, Leemann, Sessler, Haug, Pawelczyk, and Kasneci]{borisov2022deep}
Vadim Borisov, Tobias Leemann, Kathrin Sessler, Johannes Haug, Martin Pawelczyk, and Gjergji Kasneci.
\newblock Deep neural networks and tabular data: A survey.
\newblock \emph{TNNLS}, 2022.

\bibitem[Brown et~al.(2020)Brown, Mann, Ryder, Subbiah, Kaplan, Dhariwal, Neelakantan, Shyam, Sastry, Askell, et~al.]{brown2020language}
Tom Brown, Benjamin Mann, Nick Ryder, Melanie Subbiah, Jared~D Kaplan, Prafulla Dhariwal, Arvind Neelakantan, Pranav Shyam, Girish Sastry, Amanda Askell, et~al.
\newblock Language models are few-shot learners.
\newblock In \emph{NeurIPS}, 2020.

\bibitem[Cascella et~al.(2023)Cascella, Montomoli, Bellini, and Bignami]{cascella2023evaluating}
Marco Cascella, Jonathan Montomoli, Valentina Bellini, and Elena Bignami.
\newblock Evaluating the feasibility of chatgpt in healthcare: an analysis of multiple clinical and research scenarios.
\newblock \emph{Journal of Medical Systems}, 2023.

\bibitem[Chen et~al.(2022)Chen, Liao, Wan, Chen, and Wu]{chen2022danets}
Jintai Chen, Kuanlun Liao, Yao Wan, Danny~Z Chen, and Jian Wu.
\newblock {DANets}: Deep abstract networks for tabular data classification and regression.
\newblock In \emph{Proceedings of the AAAI Conference on Artificial Intelligence}, 2022.

\bibitem[Chen et~al.(2023)Chen, Liao, Fang, Chen, and Wu]{tabcaps}
Jintai Chen, KuanLun Liao, Yanwen Fang, Danny~Z. Chen, and Jian Wu.
\newblock {TabCaps}: A capsule neural network for tabular data classification with {BoW} routing.
\newblock In \emph{ICLR}, 2023.

\bibitem[Chen et~al.(2024)Chen, Yan, Chen, Chen, Wu, and Sun]{chen2024can}
Jintai Chen, Jiahuan Yan, Qiyuan Chen, Danny~Z Chen, Jian Wu, and Jimeng Sun.
\newblock Can a deep learning model be a sure bet for tabular prediction?
\newblock In \emph{SIGKDD}, pp.\  288--296, 2024.

\bibitem[Chen et~al.(2019)Chen, Liao, Chen, and Zhang]{chen2019understanding}
Pengfei Chen, Ben~Ben Liao, Guangyong Chen, and Shengyu Zhang.
\newblock Understanding and utilizing deep neural networks trained with noisy labels.
\newblock In \emph{ICML}, 2019.

\bibitem[Chen \& Guestrin(2016)Chen and Guestrin]{chen2016xgboost}
Tianqi Chen and Carlos Guestrin.
\newblock {XGBoost}: A scalable tree boosting system.
\newblock In \emph{KDD}, 2016.

\bibitem[Credit~Fusion(2011)]{credit}
Will~Cukierski Credit~Fusion.
\newblock Give me some credit, 2011.
\newblock URL \url{https://kaggle.com/competitions/GiveMeSomeCredit}.

\bibitem[Deroy et~al.(2023)Deroy, Ghosh, and Ghosh]{deroy2023ready}
Aniket Deroy, Kripabandhu Ghosh, and Saptarshi Ghosh.
\newblock How ready are pre-trained abstractive models and llms for legal case judgement summarization?
\newblock \emph{arXiv preprint arXiv:2306.01248}, 2023.

\bibitem[Detrano et~al.(1989)Detrano, Janosi, et~al.]{detrano1989international}
Robert Detrano, Andras Janosi, et~al.
\newblock International application of a new probability algorithm for the diagnosis of coronary artery disease.
\newblock \emph{The American journal of cardiology}, 1989.

\bibitem[Dinh et~al.(2022)Dinh, Zeng, Zhang, Lin, Gira, Rajput, Sohn, Papailiopoulos, and Lee]{dinh2022lift}
Tuan Dinh, Yuchen Zeng, Ruisu Zhang, Ziqian Lin, Michael Gira, Shashank Rajput, Jy-yong Sohn, Dimitris Papailiopoulos, and Kangwook Lee.
\newblock Lift: Language-interfaced fine-tuning for non-language machine learning tasks.
\newblock In \emph{NeurIPS}, 2022.

\bibitem[Fedesoriano(2020)]{fedesoriano2020stroke}
Fedesoriano.
\newblock Stroke prediction dataset.
\newblock \url{https://www.kaggle.com/datasets/fedesoriano/stroke-prediction-dataset}, 2020.

\bibitem[Gorishniy et~al.(2021)Gorishniy, Rubachev, Khrulkov, and Babenko]{gorishniy2021revisiting}
Yury Gorishniy, Ivan Rubachev, Valentin Khrulkov, and Artem Babenko.
\newblock Revisiting deep learning models for tabular data.
\newblock \emph{NeurIPS}, 2021.

\bibitem[Grinsztajn et~al.(2022)Grinsztajn, Oyallon, and Varoquaux]{grinsztajn2022tree}
Leo Grinsztajn, Edouard Oyallon, and Gael Varoquaux.
\newblock Why do tree-based models still outperform deep learning on typical tabular data?
\newblock In \emph{NeurIPS}, 2022.

\bibitem[Guo et~al.(2023)Guo, Guo, et~al.]{guo2023indeed}
Taicheng Guo, Kehan Guo, et~al.
\newblock What indeed can gpt models do in chemistry? a comprehensive benchmark on eight tasks.
\newblock \emph{arXiv preprint arXiv:2305.18365}, 2023.

\bibitem[Han et~al.(2018)Han, Yao, Yu, Niu, Xu, Hu, Tsang, and Sugiyama]{han2018co}
Bo~Han, Quanming Yao, Xingrui Yu, Gang Niu, Miao Xu, Weihua Hu, Ivor Tsang, and Masashi Sugiyama.
\newblock Co-teaching: Robust training of deep neural networks with extremely noisy labels.
\newblock In \emph{NeurIPS}, 2018.

\bibitem[Hegselmann et~al.(2023)Hegselmann, Buendia, Lang, Agrawal, Jiang, and Sontag]{hegselmann2023tabllm}
Stefan Hegselmann, Alejandro Buendia, Hunter Lang, Monica Agrawal, Xiaoyi Jiang, and David Sontag.
\newblock Tabllm: Few-shot classification of tabular data with large language models.
\newblock In \emph{AISTATS}, 2023.

\bibitem[Hemanthhari(2020)]{hemanthhari2020symptom}
Hemanthhari.
\newblock Symptoms and covid presence (may 2020 data).
\newblock \url{https://www.kaggle.com/datasets/hemanthhari/symptoms-and-covid-presence}, 2020.

\bibitem[Hoffmann et~al.(2018)Hoffmann, Bietenbeck, Lichtinghagen, and Klawonn]{hoffmann2018using}
Georg Hoffmann, Andreas Bietenbeck, Ralf Lichtinghagen, and Frank Klawonn.
\newblock Using machine learning techniques to generate laboratory diagnostic pathways—a case study.
\newblock \emph{J Lab Precis Med}, 2018.

\bibitem[Hollmann et~al.(2023)Hollmann, M{\"u}ller, Eggensperger, and Hutter]{hollmann2022tabpfn}
Noah Hollmann, Samuel M{\"u}ller, Katharina Eggensperger, and Frank Hutter.
\newblock Tabpfn: A transformer that solves small tabular classification problems in a second.
\newblock In \emph{ICLR}, 2023.

\bibitem[Huang et~al.(2019)Huang, Altosaar, and Ranganath]{huang2019clinicalbert}
Kexin Huang, Jaan Altosaar, and Rajesh Ranganath.
\newblock Clinicalbert: Modeling clinical notes and predicting hospital readmission.
\newblock \emph{arXiv preprint arXiv:1904.05342}, 2019.

\bibitem[Islam et~al.(2020)Islam, Ferdousi, Rahman, and Bushra]{islam2020likelihood}
MM~Islam, Rahatara Ferdousi, Sadikur Rahman, and Humayra~Yasmin Bushra.
\newblock Likelihood prediction of diabetes at early stage using data mining techniques.
\newblock In \emph{Computer Vision and Machine Intelligence in Medical Image Analysis}, 2020.

\bibitem[Iyyer(2019)]{churn}
Shruti Iyyer.
\newblock Churn modelling, 2019.
\newblock URL \url{https://www.kaggle.com/shrutimechlearn/churn-modelling}.

\bibitem[Johnson et~al.(2016)Johnson, Pollard, Shen, Lehman, Feng, Ghassemi, Moody, Szolovits, Anthony~Celi, and Mark]{johnson2016mimic}
Alistair~EW Johnson, Tom~J Pollard, Lu~Shen, Li-wei~H Lehman, Mengling Feng, Mohammad Ghassemi, Benjamin Moody, Peter Szolovits, Leo Anthony~Celi, and Roger~G Mark.
\newblock Mimic-iii, a freely accessible critical care database.
\newblock \emph{Scientific data}, 2016.

\bibitem[Ke et~al.(2017)Ke, Meng, Finley, Wang, Chen, Ma, Ye, and Liu]{ke2017lightgbm}
Guolin Ke, Qi~Meng, Thomas Finley, Taifeng Wang, Wei Chen, Weidong Ma, Qiwei Ye, and Tie-Yan Liu.
\newblock Lightgbm: A highly efficient gradient boosting decision tree.
\newblock In \emph{NeurIPS}, 2017.

\bibitem[Kenton \& Toutanova(2019)Kenton and Toutanova]{kenton2019bert}
Jacob Devlin Ming-Wei~Chang Kenton and Lee~Kristina Toutanova.
\newblock Bert: Pre-training of deep bidirectional transformers for language understanding.
\newblock In \emph{NAACL-HLT}, 2019.

\bibitem[Kilicarslan et~al.(2021)Kilicarslan, Celik, and Sahin]{kilicarslan2021hybrid}
Serhat Kilicarslan, Mete Celik, and {\c{S}}afak Sahin.
\newblock Hybrid models based on genetic algorithm and deep learning algorithms for nutritional anemia disease classification.
\newblock \emph{Biomedical Signal Processing and Control}, 2021.

\bibitem[Kohavi et~al.(1996)]{kohavi1996scaling}
Ron Kohavi et~al.
\newblock Scaling up the accuracy of {Naive-Bayes} classifiers: A decision-tree hybrid.
\newblock In \emph{KDD}, 1996.

\bibitem[Kojima et~al.(2022)Kojima, Gu, Reid, Matsuo, and Iwasawa]{kojima2022large}
Takeshi Kojima, Shixiang~Shane Gu, Machel Reid, Yutaka Matsuo, and Yusuke Iwasawa.
\newblock Large language models are zero-shot reasoners.
\newblock In \emph{NeurIPS}, 2022.

\bibitem[Lee et~al.(2013)]{lee2013pseudo}
Dong-Hyun Lee et~al.
\newblock Pseudo-label: The simple and efficient semi-supervised learning method for deep neural networks.
\newblock In \emph{Workshop on challenges in representation learning, ICML}, 2013.

\bibitem[Li et~al.(2019)Li, Socher, and Hoi]{li2019dividemix}
Junnan Li, Richard Socher, and Steven~CH Hoi.
\newblock Dividemix: Learning with noisy labels as semi-supervised learning.
\newblock In \emph{ICLR}, 2019.

\bibitem[Miyato et~al.(2019)Miyato, Maeda, Koyama, and Ishii]{miyato2019virtual}
Takeru Miyato, Shin-ichi Maeda, Masanori Koyama, and Shin Ishii.
\newblock Virtual adversarial training: a regularization method for supervised and semi-supervised learning.
\newblock \emph{TPAMI}, 2019.

\bibitem[Mullenbach et~al.(2018)Mullenbach, Wiegreffe, Duke, Sun, and Eisenstein]{mullenbach2018explainable}
James Mullenbach, Sarah Wiegreffe, Jon Duke, Jimeng Sun, and Jacob Eisenstein.
\newblock Explainable prediction of medical codes from clinical text.
\newblock In \emph{NAACL}, 2018.

\bibitem[Nori et~al.(2023)Nori, Lee, Zhang, Carignan, Edgar, Fusi, King, Larson, Li, Liu, et~al.]{nori2023can}
Harsha Nori, Yin~Tat Lee, Sheng Zhang, Dean Carignan, Richard Edgar, Nicolo Fusi, Nicholas King, Jonathan Larson, Yuanzhi Li, Weishung Liu, et~al.
\newblock Can generalist foundation models outcompete special-purpose tuning? case study in medicine.
\newblock \emph{arXiv preprint arXiv:2311.16452}, 2023.

\bibitem[O'Donnell \& Elosua(2008)O'Donnell and Elosua]{o2008cardiovascular}
Christopher~J O'Donnell and Roberto Elosua.
\newblock Cardiovascular risk factors. insights from framingham heart study.
\newblock \emph{Revista Espa{\~n}ola de Cardiolog{\'\i}a (English Edition)}, 2008.

\bibitem[OpenAI(2022)]{openai2022gpt3}
OpenAI.
\newblock Openai: Introducing chatgpt.
\newblock \url{https://openai.com/blog/chatgpt}, 2022.

\bibitem[OpenAI(2023)]{openai2023gpt4}
OpenAI.
\newblock Gpt-4 technical report.
\newblock \url{https://cdn.openai.com/papers/gpt-4.pdf}, 2023.

\bibitem[Qian et~al.(2023)Qian, Wang, Li, Li, and Yan]{qian2023limitations}
Jing Qian, Hong Wang, Zekun Li, Shiyang Li, and Xifeng Yan.
\newblock Limitations of language models in arithmetic and symbolic induction.
\newblock In \emph{ACL}, 2023.

\bibitem[Qin et~al.(2023)Qin, Liang, et~al.]{qin2023toolllm}
Yujia Qin, Shihao Liang, et~al.
\newblock Toolllm: Facilitating large language models to master 16000+ real-world apis.
\newblock \emph{arXiv preprint arXiv:2307.16789}, 2023.

\bibitem[Radford et~al.(2019)Radford, Wu, Child, Luan, Amodei, Sutskever, et~al.]{radford2019language}
Alec Radford, Jeffrey Wu, Rewon Child, David Luan, Dario Amodei, Ilya Sutskever, et~al.
\newblock Language models are unsupervised multitask learners.
\newblock \emph{OpenAI blog}, 2019.

\bibitem[Ramana et~al.(2012)Ramana, Babu, and Venkateswarlu]{ramana2012critical}
Bendi~Venkata Ramana, M~Surendra~Prasad Babu, and NB~Venkateswarlu.
\newblock A critical comparative study of liver patients from usa and india: an exploratory analysis.
\newblock \emph{IJCSI}, 2012.

\bibitem[Smith et~al.(1988)Smith, Everhart, et~al.]{smith1988using}
Jack~W Smith, James~E Everhart, et~al.
\newblock Using the adap learning algorithm to forecast the onset of diabetes mellitus.
\newblock In \emph{Proceedings of the annual symposium on computer application in medical care}, 1988.

\bibitem[Song et~al.(2022)Song, Kim, Park, Shin, and Lee]{song2022learning}
Hwanjun Song, Minseok Kim, Dongmin Park, Yooju Shin, and Jae-Gil Lee.
\newblock Learning from noisy labels with deep neural networks: A survey.
\newblock \emph{IEEE Transactions on Neural Networks and Learning Systems}, 2022.

\bibitem[Tarvainen \& Valpola(2017)Tarvainen and Valpola]{tarvainen2017mean}
Antti Tarvainen and Harri Valpola.
\newblock Mean teachers are better role models: Weight-averaged consistency targets improve semi-supervised deep learning results.
\newblock \emph{NeurIPS}, 2017.

\bibitem[Wang et~al.(2022)Wang, Roberts, Hesslow, Le~Scao, Chung, Beltagy, Launay, and Raffel]{wang2022language}
Thomas Wang, Adam Roberts, Daniel Hesslow, Teven Le~Scao, Hyung~Won Chung, Iz~Beltagy, Julien Launay, and Colin Raffel.
\newblock What language model architecture and pretraining objective works best for zero-shot generalization?
\newblock In \emph{ICML}, 2022.

\bibitem[Wang et~al.(2023)Wang, Wei, Schuurmans, Le, Chi, Narang, Chowdhery, and Zhou]{wang2022self}
Xuezhi Wang, Jason Wei, Dale Schuurmans, Quoc~V Le, Ed~H Chi, Sharan Narang, Aakanksha Chowdhery, and Denny Zhou.
\newblock Self-consistency improves chain of thought reasoning in language models.
\newblock In \emph{ICLR}, 2023.

\bibitem[Wang \& Sun(2022)Wang and Sun]{wang2022transtab}
Zifeng Wang and Jimeng Sun.
\newblock Transtab: Learning transferable tabular transformers across tables.
\newblock In \emph{NeurIPS}, 2022.

\bibitem[Wei et~al.(2021)Wei, Bosma, Zhao, Guu, Yu, Lester, Du, Dai, and Le]{wei2021finetuned}
Jason Wei, Maarten Bosma, Vincent Zhao, Kelvin Guu, Adams~Wei Yu, Brian Lester, Nan Du, Andrew~M Dai, and Quoc~V Le.
\newblock Finetuned language models are zero-shot learners.
\newblock In \emph{ICLR}, 2021.

\bibitem[Wei et~al.(2022)Wei, Wang, Schuurmans, Bosma, Xia, Chi, Le, Zhou, et~al.]{wei2022chain}
Jason Wei, Xuezhi Wang, Dale Schuurmans, Maarten Bosma, Fei Xia, Ed~Chi, Quoc~V Le, Denny Zhou, et~al.
\newblock Chain-of-thought prompting elicits reasoning in large language models.
\newblock \emph{NeurIPS}, 2022.

\bibitem[Wen et~al.(2024)Wen, Zhang, Zheng, Xu, and Bian]{wen2024supervised}
Xumeng Wen, Han Zhang, Shun Zheng, Wei Xu, and Jiang Bian.
\newblock From supervised to generative: A novel paradigm for tabular deep learning with large language models.
\newblock In \emph{SIGKDD}, pp.\  3323--3333, 2024.

\bibitem[Yan et~al.(2023)Yan, Chen, Wu, Chen, and Wu]{yan2023t2g}
Jiahuan Yan, Jintai Chen, Yixuan Wu, Danny~Z Chen, and Jian Wu.
\newblock {T2G-Former}: Organizing tabular features into relation graphs promotes heterogeneous feature interaction.
\newblock In \emph{AAAI}, 2023.

\bibitem[Yan et~al.(2024{\natexlab{a}})Yan, Chen, Wang, Chen, and Wu]{yan2024team}
Jiahuan Yan, Jintai Chen, Qianxing Wang, Danny~Z Chen, and Jian Wu.
\newblock Team up gbdts and dnns: Advancing efficient and effective tabular prediction with tree-hybrid mlps.
\newblock In \emph{SIGKDD}, pp.\  3679--3689, 2024{\natexlab{a}}.

\bibitem[Yan et~al.(2024{\natexlab{b}})Yan, Zheng, Xu, Zhu, Chen, Sun, Wu, and Chen]{yan2024making}
Jiahuan Yan, Bo~Zheng, Hongxia Xu, Yiheng Zhu, Danny Chen, Jimeng Sun, Jian Wu, and Jintai Chen.
\newblock Making pre-trained language models great on tabular prediction.
\newblock In \emph{ICLR}, 2024{\natexlab{b}}.

\bibitem[Yao et~al.(2023)Yao, Yu, Zhao, Shafran, Griffiths, Cao, and Narasimhan]{yao2023tree}
Shunyu Yao, Dian Yu, Jeffrey Zhao, Izhak Shafran, Tom Griffiths, Yuan Cao, and Karthik Narasimhan.
\newblock Tree of thoughts: Deliberate problem solving with large language models.
\newblock In \emph{NeurIPS}, 2023.

\bibitem[Zan et~al.(2023)Zan, Chen, Zhang, Lu, Wu, Guan, Yongji, and Lou]{zan2023large}
Daoguang Zan, Bei Chen, Fengji Zhang, Dianjie Lu, Bingchao Wu, Bei Guan, Wang Yongji, and Jian-Guang Lou.
\newblock Large language models meet nl2code: A survey.
\newblock In \emph{ACL}, 2023.

\bibitem[Zhang et~al.(2024{\natexlab{a}})Zhang, Yu, Li, Dong, Su, Chu, and Yu]{zhang2024mm}
Duzhen Zhang, Yahan Yu, Chenxing Li, Jiahua Dong, Dan Su, Chenhui Chu, and Dong Yu.
\newblock Mm-llms: Recent advances in multimodal large language models.
\newblock \emph{arXiv preprint arXiv:2401.13601}, 2024{\natexlab{a}}.

\bibitem[Zhang et~al.(2024{\natexlab{b}})Zhang, Huang, Jin, and Lu]{zhang2024vision}
Jingyi Zhang, Jiaxing Huang, Sheng Jin, and Shijian Lu.
\newblock Vision-language models for vision tasks: A survey.
\newblock \emph{IEEE Transactions on Pattern Analysis and Machine Intelligence}, 2024{\natexlab{b}}.

\bibitem[Zhang et~al.(2023{\natexlab{a}})Zhang, Zhang, Li, and Smola]{zhang2022automatic}
Zhuosheng Zhang, Aston Zhang, Mu~Li, and Alex Smola.
\newblock Automatic chain of thought prompting in large language models.
\newblock In \emph{ICLR}, 2023{\natexlab{a}}.

\bibitem[Zhang et~al.(2023{\natexlab{b}})Zhang, Zhang, Li, Zhao, Karypis, and Smola]{zhang2023multimodal}
Zhuosheng Zhang, Aston Zhang, Mu~Li, Hai Zhao, George Karypis, and Alex Smola.
\newblock Multimodal chain-of-thought reasoning in language models.
\newblock \emph{arXiv preprint arXiv:2302.00923}, 2023{\natexlab{b}}.

\bibitem[Zhao et~al.(2023)Zhao, Zhou, et~al.]{zhao2023survey}
Wayne~Xin Zhao, Kun Zhou, et~al.
\newblock A survey of large language models.
\newblock \emph{arXiv preprint arXiv:2303.18223}, 2023.

\end{thebibliography}
\bibliographystyle{iclr2025_conference}

\appendix
\newpage

\section{Limitations \& Impacts} \label{limitation}

As discussed in Sec.~\ref{sec:other-domain}, though our \texttt{SERSAL} is distinguished from traditional prompting methods by its non-linguistic mechanism, it still requires the LLMs with latent knowledge in the target domain to be effective. Therefore, in practice the user should have prior understanding of the used LLM's capability or advantageous application fields. \texttt{SERSAL} contributes to the progress in both LLM prompting and tabular data community through providing a novel interface to adapt untapped knowledge in LLMs to the tabular prediction tasks in a zero-shot manner, which is particularly useful in the regime where limited data or annotation is available.

\section{Datasets and Experiment Details} \label{app:data-info}

We provide detailed data information of the experiment tabular datasets in Table~\ref{tab:med-tables}. We drop the samples with missing features and adopt the same preprocessing as ~\cite{gorishniy2021revisiting} before training. For MIMIC-III discharge summary dataset~\citep{johnson2016mimic,mullenbach2018explainable} used in Fig.~\ref{main-fig}(a), we retain the most frequent 5 labels (medical codes) since our goal is just to demonstrate the prompting effectiveness on medical textual tasks and conducting validation on the full label version (several thousands labels) is inconvenient. During conducting zero-shot prompting for GPT-3.5v and GPT-4v on the MIMIC-III dataset, we follow the PhysioNet Credentialed Data Use Agreement~\footnote{\url{https://physionet.org/news/post/415}} and enroll in the Azure OpenAI service without human review of the data to protect the data from third-party access.

\begin{table*}[ht]
    \setlength{\tabcolsep}{0.15em}
    \centering
    \resizebox{\textwidth}{!}{
    \begin{tabular}{l|l|l|l|c|l}
    \hline
    Dataset                          & Abbr. & \# Sample & \# Feature & P/N                  & Source Link                                                                                      \\ \hline
    Indian Liver Patient Records     & LI    & 583       & 10         & 2.51                      & \url{https://www.kaggle.com/datasets/uciml/indian-liver-patient-records}                           \\
    Pima Indians Diabetes Database   & PID   & 768       & 8          & 0.54                      & \url{https://www.kaggle.com/datasets/uciml/pima-indians-diabetes-database}                         \\
    Framingham Heart Study           & FH    & 4238      & 15         & 0.18                      & \url{https://www.kaggle.com/datasets/mohannapd/ramingham-heart-study} \\
    Stroke Prediction                & ST    & 5110      & 7          & 0.04                      & \url{https://www.kaggle.com/datasets/fedesoriano/stroke-prediction-dataset}                        \\
    Hepatitis C Prediction           & HE    & 615       & 12         & 0.11                      & \url{https://www.kaggle.com/datasets/fedesoriano/hepatitis-c-dataset}                              \\
    COVID-19                         & CO    & 5434      & 20         & 4.17                      & \url{https://www.kaggle.com/datasets/hemanthhari/symptoms-and-covid-presence}                      \\
    Lung Cancer Prediction           & LC    & 309       & 15         & 6.92                      & \url{https://www.kaggle.com/datasets/mysarahmadbhat/lung-cancer}                                   \\
    Heart Failure Prediction         & HF    & 303       & 13         & 0.80                      & \url{https://archive.ics.uci.edu/dataset/45/heart+disease}                                         \\
    Early Classification of Diabetes & ECD   & 520       & 16         & 1.60                      & \url{https://www.kaggle.com/datasets/andrewmvd/early-diabetes-classification}                      \\
    Anemia Disease                   & AN    & 15300     & 24         & 0.57                      & \url{https://www.kaggle.com/datasets/serhathoca/anemia-disease}                                    \\
    Churn Modeling                   & -     & 10000     & 10         & 0.26     & -                                                                                                \\
    Give Me Some Credit              & -     & 16714     & 10         & 1.00     & \url{https://www.kaggle.com/c/GiveMeSomeCredit}                                                        \\
    US Adult Income                  & -     & 48842     & 14         & 0.31     & \url{https://www.kaggle.com/datasets/johnolafenwa/us-census-data}                                      \\ \hline
    \end{tabular}}
    \caption{Detailed data information of used tabular datasets (10 from the medical domain and 3 from others). ``P/N'' denotes the amount ratio of positive samples and negative ones.}
    \label{tab:med-tables}
    \vskip -0.8 em
\end{table*}

\section{Results on Clinical Trial Datasets}

We evaluate \texttt{SERSAL} on clinical trail mortality datasets, which require specialized scientific knowledge for clinical trials. Although \texttt{SERSAL} prompting with GPT-3.5 cannot directly achieve good performance on such vertical tasks, further performance gains are still observed once we use more powerful GPT-4, indicating room for continuous improvement as more advanced LLMs appear.

\begin{table*}[th]
\centering
\resizebox{\textwidth}{!}{
\begin{tabular}{@{}l|ccccc@{}}
\toprule
       & N00041119 & N00174655 & N00312208 & N00079274 & N00694382    \\ \midrule
FSSM$^\ast$(supervised FT-T)   & 62.38 & 89.20 & 77.83 & 71.78 & 73.89 \\ \midrule
0-shot (GPT-3.5) & 56.79 & 73.08 & 63.49 & 59.85 & 62.70 \\
CoT (GPT-3.5) & 56.79 & 73.08 & 60.73 & 59.85 & 62.70 \\
\texttt{SERSAL} (GPT-3.5) & \textbf{58.31} & \textbf{82.64} & \textbf{71.92} & \textbf{64.17} & \textbf{66.31}
 \\ \midrule       
\texttt{SERSAL} (GPT-4) & 65.08 & 88.62 & 78.39 & 67.94 & 71.47 \\ \bottomrule
\end{tabular}}
\caption{The AUC scores (\%) of different tabular prediction schemes on clinical trail mortality datasets used in ~\cite{wang2022transtab} (see \href{https://clinicaltrials.gov/}{ClinicalTrials.gov}). The similar denotations are used as Table~\ref{main-res}. No gold labels are used for prompting methods here. It can be seen \texttt{SERSAL} can achieve continuous improvement and even perform comparably with the traditional supervised paradigm once more powerful base LLMs are applied.} 
\label{clinical-trial}
\end{table*}

\section{Mechanism Explanation of DivideMix in SERSAL}

To make the paper friendly to the audiences from different background, in this section we provide detailed mechanism explanation of learning with noisy labels (LNL) and how to learn a better small (neural network) model from LLM noisy annotations using DivideMix.

\paragraph{DivideMix mechanism in SERSAL}

In the traditional noisy data learning field, it was theoretically proved and empirically observed that the “memorization” behavior of neural networks leads to different optimization behavior on real data and noisy ones that neural networks tend to learn simple patterns first before fitting label noise~\citep{arpit2017closer}. Based on this theoretical foundation, a typical group of LNL methods~\citep{berthelot2019mixmatch,li2019dividemix} exploit per-sample training loss to judge the noisy labels, for example, in our paper we adopt DivideMix~\citep{li2019dividemix} to learn a small model using LLM noisy annotations, which models the noise probabilities of each sample by dynamically fitting a Gaussian Mixture Model (GMM) on per-sample losses, all training samples are divided into a clean set and a noisy set based on a probability threshold $\tau$. During the DivideMix training process, samples in the clean set are used for supervised learning (using their soft LLM annotations), while ones in the noisy set is used in an unsupervised manner (only using their features), e.g., learn with regularization loss or reconstruction task. The process will be ended until the average loss of heuristically selected early stopping subset (high-LLM-confidence samples $D_{es}$ in Algorithm 1) is converged, i.e., the loss of early stopping subset is not decreased for $m$ epochs. Notably, clean sample is not equivalent to high-LLM-confidence sample, but the sample which LLM annotation is easier to fit by the small tabular model. Since the small model (i.e., FT-Transformer here) is only supervised by clean data and regularized on noisy data, all data is sufficiently and reasonably exploited to acquire a better pattern.

\paragraph{DivideMix hyperparameters in SERSAL}

We refer to the original hyperparameter settings in DivideMix paper [4] and only search the temperature ($T$) in $\left \{\text{0.5}, \text{5.0}, \text{10.0}\right \}$, with fixed regularization loss weight $L_u$ to 25, clean probability $\tau$ to 0.9, and the learning rate of the small model (FT-Transformer) to 1e-4. Additionally, we uniformly introduce the early stopping patience $m$ to 5. The best temperature is selected based on the training loss of early stopping subset $D_{es}$.

\begin{figure*}[ht]
\centering
\includegraphics[width=1.0\linewidth]{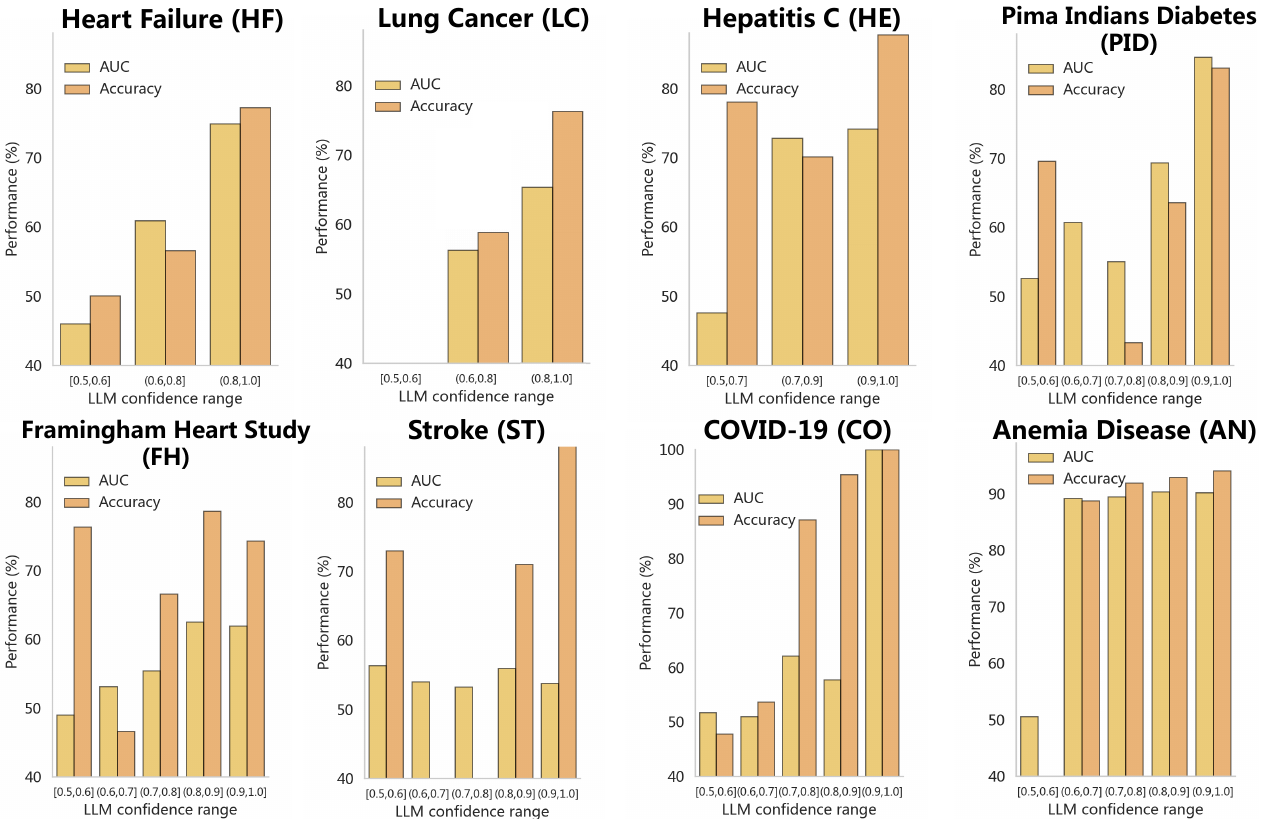}
\caption{Performances in different LLM confidence ranges on other eight datasets. The overall trend of high-confidence samples being relatively more reliable still holds.}
\label{conf-perform2}
\end{figure*}

\begin{table*}[th]
\centering
\resizebox{\textwidth}{!}{
\begin{tabular}{@{}l|cccccccccc@{}}
\toprule
                   & HF    & LC    & ECD   & LI    & HE    & PID   & FH    & ST    & CO    & AN    \\ \midrule
0-shot GPT-3.5 \#1 & 71.88 & 78.87 & 85.71 & 76.81 & 68.51 & 73.12 & 60.32 & 63.01 & 82.60 & 90.43 \\
SERSAL \#1         & 91.39 & 85.42 & 86.40 & 79.39 & 85.14 & 78.97 & 63.97 & 76.36 & 96.85 & 98.37 \\ \midrule
0-shot GPT-3.5 \#2 & 87.58 & 83.74 & 86.42 & 80.26 & 86.18 & 79.26 & 63.86 & 73.62 & 91.29 & 93.62 \\
SERSAL \#2         & 92.03 & 86.15 & 87.00 & 82.47 & 87.32 & 80.61 & 65.27 & 79.58 & 97.20 & 98.93 \\ \midrule
0-shot GPT-3.5 \#3 & 89.26 & 85.39 & 87.81 & 82.91 & 86.87 & 81.47 & 64.12 & 76.37 & 93.65 & 94.13 \\
SERSAL \#3         & 93.58 & 85.42 & 89.00 & 84.07 & 89.57 & 81.83 & 65.27 & 80.93 & 97.02 & 98.60 \\ \bottomrule
\end{tabular}}
\caption{The AUC score variation of \texttt{SERSAL} outputs and zero-shot prompting of the tuned GPT-3.5 on all datasets from Table~\ref{main-res} during three loops.}
\label{sersal-iter2}
\end{table*}

\end{document}